\documentclass[11pt,a4paper]{article}
\usepackage[utf8]{inputenc}
\usepackage[T1]{fontenc}

\usepackage[hyperref]{acl2017}
\usepackage{times}
\usepackage{latexsym}

\usepackage{url}
\usepackage{siunitx}
\usepackage{tabularx}

\usepackage{amsmath}

\usepackage{todonotes}
\usepackage{enumitem}
\usepackage{booktabs}

\usepackage{amssymb}
\usepackage{pifont}
\usepackage{fontawesome}
\newcommand{\cmark}{\ding{51}}%
\newcommand{\xmark}{\ding{55}}%

\DeclareMathOperator*{\argmax}{arg\,max}

\newcommand*{\numTwo}[1] {\num[round-precision=2]{#1}}
\newcommand*{\numOne}[1] {\num[round-precision=1]{#1}}
\newcommand*{\everymodeprime}{\ensuremath{\prime}}

\definecolor{lincolngreen}{rgb}{0.11, 0.35, 0.02}
\definecolor{brickred}{rgb}{0.8, 0.25, 0.33}
\definecolor{brass}{rgb}{0.71, 0.65, 0.26}

\newcommand{\correctmark}{{\color{lincolngreen}{\cmark}}}
\newcommand{\incorrectmark}{\color{brickred}{\xmark}}
\newcommand{\questionmark}{\color{brass}{\faQuestion}}

\aclfinalcopy 

\title{Beyond {E}nglish-Only Reading Comprehension: \\ Experiments in Zero-Shot Multilingual Transfer for {B}ulgarian}

\author{Momchil Hardalov$^1$ \quad Ivan Koychev$^1$ \quad Preslav Nakov$^2$ \\
  $^1$Sofia University ``St. Kliment Ohridski'', Bulgaria, \\
  $^2$Qatar Computing Research Institute, HBKU, Qatar, \\
  {\tt \{hardalov, koychev\}@fmi.uni-sofia.bg} \\
  {\tt pnakov@hbku.edu.qa}
}

\date{}

\begin{document}

\maketitle

\begin{abstract}
Recently, reading comprehension models achieved near-human performance on large-scale datasets 
such as SQuAD, CoQA, MS Macro, RACE, etc. This is largely due to the release of pre-trained contextualized representations such as BERT and ELMo, which can be  fine-tuned for the target task. Despite those advances and the creation of more challenging datasets, most of the work is still done for English.
Here, we study the effectiveness of multilingual BERT fine-tuned on large-scale English datasets for reading comprehension (e.g., for RACE), and we apply it to Bulgarian multiple-choice reading comprehension. We propose a new 
dataset containing 2,221 questions from matriculation exams for twelfth grade in various subjects ---history, biology, geography 
and philosophy---, and 412 additional questions from online quizzes in history. While the quiz authors gave no relevant context, we incorporate knowledge from Wikipedia, retrieving documents matching the combination 
of question + each answer option. Moreover, we experiment with different indexing and pre-training strategies. 
The evaluation results show accuracy of 42.23\%, which is well above the baseline of 24.89\%.

\end{abstract}

\section{Introduction}
\label{sec:introduction}

The ability to answer questions is natural to humans, independently of their native language, and, once learned, it can be easily transferred to another language. After understanding the question, we typically depend on our background knowledge, and on relevant information from external sources. 

\noindent Machines do not have the reasoning ability of humans, but they are still able to learn concepts. The growing interest in teaching machines to answer questions posed in natural language has led to the introduction of various new datasets for different tasks such as reading comprehension, both extractive, e.g., span-based ~\citep{nguyen2016ms,trischler-etal-2017-newsqa,joshi-etal-2017-triviaqa,rajpurkar-etal-2018-know,reddy2018coqa}, and non-extractive, e.g., multiple-choice questions \citep{richardson-etal-2013-mctest,lai-etal-2017-race,clark2018think,mihaylov-etal-2018-suit,sundream2019}. Recent advances in neural network architectures, especially the raise of the Transformer~\citep{NIPS2017_7181:transformer}, and better contextualization of language models~\citep{Peters:2018:ELMo,devlin2018bert,radford2018improving,grave-etal-2018-learning,howard-ruder-2018-universal,radford2019language,yang2019xlnet,dai2019transformer} offered new opportunities to advance the field. 

Here, we investigate skill transfer from a high-resource language, i.e.,~English, to a low-resource one, i.e.,~Bulgarian, for the task of multiple-choice reading comprehension. Most previous work \citep{DBLP:journals/corr/abs-1902-00993, radford2018improving,DBLP:journals/corr/abs-1803-09074, sun-etal-2019-improving} was monolingual, and a relevant context for each question was available a priori. We take the task a step further by exploring the capability of a neural comprehension model in a multilingual setting using external commonsense knowledge. Our approach is based on the multilingual cased BERT~\citep{devlin2018bert} fine-tuned on the RACE  dataset~\citep{lai-etal-2017-race}, which contains over 87,000 English multiple-choice school-level science questions. For evaluation, we build a novel dataset for Bulgarian. 
We further experiment with pre-training the model over stratified Slavic corpora in Bulgarian, Czech, and Polish Wikipedia articles, and Russian news, as well as with various document retrieval strategies. 

\noindent Finally, we address the resource scarceness in low-resource languages and the absence of question contexts in our dataset by extracting relevant passages from Wikipedia articles.\\

Our contributions are as follows: 
\begin{itemize}
     \item We introduce a new dataset for reading comprehension in a low-resource language such as Bulgarian.  The dataset contains a total of 2,636 multiple-choice questions without contexts from matriculation exams and online quizzes. These questions cover a large variety of science topics in biology, philosophy, geography, and history.
     \item We study the effectiveness of zero-shot transfer from English to Bulgarian for the task of multiple-choice reading comprehension, using Multilingual and Slavic BERT~\citep{devlin2018bert}, fine-tuned on large corpora, such as RACE~\citep{lai-etal-2017-race}.
     \item We design a general-purpose pipeline\footnote{The dataset and the source code are available at \url{http://github.com/mhardalov/bg-reason-BERT}} for extracting relevant contexts from an external corpus of unstructured documents using information retrieval.
 \end{itemize}

The rest of this paper is organized as follows: The next section presents related work. Section~\ref{sec:approach} describes our approach. Details about the newly-proposed multiple-choice Bulgarian dataset are given in Section~\ref{sec:data}. All experiments are described in Section~\ref{sec:experiments}. Finally, Section~\ref{sec:conclustions} concludes and points to possible directions for future work. 
    \section{Related Work}
\label{sec:relatedwork}

\subsection{Machine Reading Comprehension}

The growing interest in machine reading comprehension (MRC) has led to the release of various datasets for both extractive \citep{nguyen2016ms,trischler-etal-2017-newsqa,joshi-etal-2017-triviaqa,rajpurkar-etal-2018-know,reddy2018coqa} and non-extractive \citep{richardson-etal-2013-mctest,Penas_overviewof2014,lai-etal-2017-race,clark2018think,mihaylov-etal-2018-suit,sundream2019} comprehension. Our work primarily focuses on the non-extractive multiple-choice type, designed by educational experts, since their task is very close to our newly-proposed dataset, and are expected to be well-structured and error-free \citep{sundream2019}.

\noindent These datasets brought a variety of models and approaches. The usage of external knowledge has been an interesting topic,  e.g.,~\citet{Chen:2017:DrQA} used Wikipedia knowledge for answering open-domain questions,~\citet{DBLP:journals/corr/abs-1902-00993} applied entity discovery and linking as a source of prior knowledge. \citet{sun-etal-2019-improving} explored different reading strategies such as back and forth reading, highlighting, and self-assessment. \citet{ni-etal-2019-learning} focused on finding essential terms and removing distraction words, followed by reformulation of the question, in order to find better evidence before sending a query to the MRC system. A simpler approach was presented by \citet{clark2016combining}, who leveraged information retrieval, corpus statistics, and simple inference over a semi-automatically constructed knowledge base for answering fourth-grade science questions. 

Current state-of-the-art approaches in machine reading comprehension are grounded on transfer learning and fine-tuning of language models \citep{Peters:2018:ELMo,conneau-etal-2018-xnli,devlin2018bert}. \citet{yang-etal-2019-end} presented an open-domain extractive reader based on BERT~\citep{devlin2018bert}. \citet{radford2018improving} used generative pre-training of a Transformer~\citep{NIPS2017_7181:transformer} as a language model, transferring it to downstream tasks such as natural language understanding, reading comprehension, etc. 

Finally, there has been a Bulgarian MRC dataset ~\citep{Penas_overviewof}.  It was used by \citet{simov2012bulgarian}, who converted the question-answer pairs to declarative sentences, and measured their similarity to the context, transforming both to a bag of linguistic units: lemmata, POS tags, and dependency relations.

\subsection{(Zero-Shot) Multilingual Models}

Multilingual embeddings helped researchers to achieve new state-of-the-art results on many NLP tasks. While many pre-trained model~\citep{grave-etal-2018-learning,devlin2018bert,lample2019cross} are available, the need for task-specific data in the target language still remains. Learning such models is language-independent, and representations for common words remain close in the latent vector space for a single language, albeit unrelated for different languages. A possible approach to overcome this effect is to learn an alignment function between spaces~\citep{DBLP:journals/corr/abs-1812-10464,joty-etal-2017-cross}.

\noindent Moreover, zero-shot application of fine-tuned multilingual language models \citep{devlin2018bert,lample2019cross} on XNLI~\citep{conneau-etal-2018-xnli}, a corpus containing sentence pairs annotated with textual entailment and translated into 14 languages, has shown very close results to such by a language-specific model.

Zero-shot transfer and multilingual models had been a hot topic in (neural) machine translation (MT) in the past several years. \citet{johnson-etal-2017-googles} introduced a simple tweak to a standard sequence-to-sequence \citep{sutskever2014sequence} model by adding a special token to the encoder's input, denoting the target language, allowing a zero-shot learning for new language pairs. Recent work in zero-resource translation outlined different strategies for learning to translate without having a parallel corpus between the two target languages. First, a many-to-one approach was adopted by \citet{firat-etal-2016-zero} based on building a corpus from a single language paired with many others, allowing simultaneous training of multiple models, with a shared attention layer. A many-to-many relationship between languages was later used by \citet{aharoni-etal-2019-massively}, in an attempt to train a single Transformer~\citep{NIPS2017_7181:transformer} model. 

Pivot-language approaches can also be used to overcome the lack of parallel corpora for the source--target language pair. \citet{chen-etal-2017-teacher} used a student-teacher framework to train an NMT model, using a third language as a pivot. A similar idea was applied to MRC by \citet{asai2018multilingual}, who translated each question to a pivot language, and then found the correct answer in the target language using soft-alignment attention scores.
\section{Model}
\label{sec:approach}

Our model has three components: (\textit{i})~a context retrieval module, which tries to find good explanatory passages for each question-answer pair, from a corpus of non-English documents, as described in Section~\ref{sec:ctxretriever}, (\textit{ii})~a multiple-choice reading comprehension module pre-trained on English data and then applied to the target language in a zero-shot fashion, i.e., without further training or additional fine-tuning, to a target (non-English) language, as described in Section~\ref{sec:bertformc}, and (\textit{iii})~a voting mechanism, described in Section~\ref{sec:selstrategies}, which combines multiple passages from (\textit{i}) and their scores from (\textit{ii}) in order to obtain a single (most probable) answer for the target question.

\subsection{Context Retriever}
\label{sec:ctxretriever}

Most public datasets for reading comprehension \citep{richardson-etal-2013-mctest,lai-etal-2017-race,sundream2019,rajpurkar-etal-2018-know,reddy2018coqa,mihaylov-etal-2018-suit} contain not only questions with possible answers, but also an evidence passage for each question. This limits the task to question answering over a piece of text, while an open-domain scenario is much more challenging and much more realistic. 
Moreover, a context in which the answer can be found is not easy to retrieve, sometimes even for a domain expert. Finally, data scarceness in low-resource languages poses further challenges for finding resources and annotators.

In order to enable search for appropriate passages for non-English questions, we created an inverted index from Wikipedia articles using Elasticsearch.\footnote{\url{http://www.elastic.co/}}
We used the original dumps for the entire Wikipage,\footnote{\url{http://dumps.wikimedia.org/}} and we preprocessed the data leaving only plain textual content, e.g.,~removing links, HTML tags, tables, etc. Moreover, we split the article's body using two strategies: a sliding window and a paragraph-based approach. Each text piece with its corresponding article title was processed by applying word-based tokenization, lowercasing, stop-words removal, stemming~\citep{Nakov:2003:BIS:973620.973690,savoy2007searching}, and $n$-gram extraction. Finally, the matching between a question and a passage was done using cosine similarity and BM25~\cite{Robertson:2009:PRF:1704809.1704810}.

\subsection{BERT for Multiple-Choice RC}
\label{sec:bertformc}

The recently-proposed BERT~\citep{devlin2018bert} framework is applicable to a vast number of NLP tasks. A shared characteristic between all of them is the form of the input sequences: a single sentence or a pair of sentences separated by the [SEP] special token, and a classification token ([CLS]) added at the beginning of each example. In contrast, the input for multiple-choice reading comprehension questions is assembled by three sentence pieces, i.e.,~context passage, question, and possible answer(s). Our model follows a simple strategy of concatenating the option (candidate answer) at the end of a question. Following the notation of~\citet{devlin2018bert}, the input sequence can be written as follows:

\medskip
\noindent
\emph{[CLS] Passage [SEP] Question + Option [SEP]}
\medskip

\begin{figure}[t]
    \centering
    \includegraphics[width=\columnwidth]{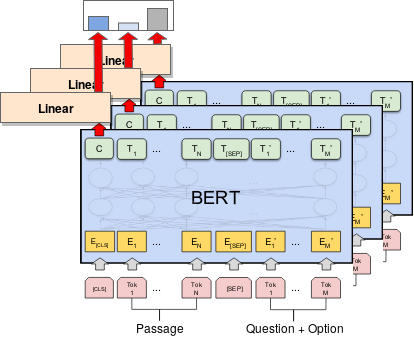}
    \caption{BERT for multiple-choice reasoning.}
    \label{fig:bert_model}
    \vspace{-0.2in}
\end{figure}

As recommended by \citet{devlin2018bert}, we introduce a new task-specific parameter vector $L$, $L \in \mathbb{R}^H$, where $H$ is the hidden size of the model. In order to obtain a score for each passage-question-answer triplet, we take the dot product between $L$ and the final hidden vector for the classification token ([CLS]), thus ending up with $N$ unbounded numbers: one for each option. Finally, we normalize the scores by adding a softmax layer, as shown in Figure~\ref{fig:bert_model}. During fine-tuning, we optimize the model's parameters by maximizing the log-probability of the correct answer.

\subsection{Answer Selection Strategies}
\label{sec:selstrategies}

Finding evidence passages that contain information about the correct answer is crucial for reading comprehension systems. The context retriever may be extremely sensitive to the formulation of a question. The latter can be very general, or can contain insignificant rare words, which can bias the search. Thus, instead of using only the first-hit document, we should also evaluate lower-ranked ones. Moreover, knowing the answer candidates can enrich the search query, resulting in improved, more answer-oriented passages. This approach leaves us with a set of contexts that need to be evaluated by the MRC model in order to choose a single correct answer. Prior work suggests several different strategies: \citet{Chen:2017:DrQA} used the raw predicted probability from a recurrent neural network (RNN), \citet{yang-etal-2019-end} tuned a hyper-parameter to balance between the retriever score and the reading model's output, while \citet{DBLP:journals/corr/abs-1902-00993} and \citet{ni-etal-2019-learning} concatenated the results from sentence-based retrieval into a single contextual passage. 

\noindent In our experiments below, we adopt a simple summing strategy. We evaluate each result from the context retriever against the question and the possible options (see Section~\ref{sec:bertformc} for more details), thus obtaining a list of raw probabilities. We found empirically that explanatory contexts assign higher probability to the related answer, while general or uninformative passages lead to stratification of the probability distribution over the answer options. We formulate this as follows:
\begin{equation}
\   \small
    Pr(a_j|p;q) = \frac{exp(BERT(p,q+a_j))}{\sum_{j\everymodeprime}{exp(BERT(p,q+a_j\everymodeprime))}},
    \label{eq:bertsoftmax}
    \vspace{-0.1in}
\end{equation}{}

\noindent where $p$ is a passage, $q$ is a question, $A$ is the set of answer candidates, and $a_j \in A$.

We select the final answer as follows:
\begin{equation}
\small
Ans = \argmax_{a \in A} \sum_{p \in P}{Pr(A|p;q)}
\label{eq:anssel}
\end{equation}

\section{Data}
\label{sec:data}

Our goal is to build a task for a low-resource language, such as Bulgarian, as close as possible to the multiple-choice reading comprehension setup for high-resource languages such as English. This will allow us to evaluate the limitations of transfer learning in a multilingual setting. One of the largest datasets for this task is RACE~\citep{lai-etal-2017-race}, with a total of 87,866 training questions with four answer candidates for each. Moreover, there are 25,137 contexts mapped to the questions and their correct answers.

\begin{table*}[t]
    \centering
    \begin{tabular}{lrcccc}
        \toprule
        \bf Domain & \bf \#QA-pairs & \bf \#Choices & \bf Len Question & \bf Len Options & \bf Vocabulary Size \\
        \hline
        \multicolumn{6}{c}{12th Grade Matriculation Exam} \\
        \hline
        Biology & 437 & 4 & \numOne{10.437070938215102} & \numOne{2.643592677345538} & $2,414$ ($12,922$)  \\
        Philosophy & 630 & 4 & \numOne{8.90952380952381} & \numOne{2.938888888888889}  &  $3,636$  ($20,392$) \\
        Geography & 612 & 4 & \numOne{12.830065359477125} & \numOne{2.46609477124183} & $3,239$ ($17,668$) \\
        History & 542 & 4 & \numOne{23.73761467889908} & \numOne{3.6440366972477065} & $5,466$ ($20,456$) \\
        \hline
        \multicolumn{6}{c}{Online History Quizzes} \\
        \hline
        Bulgarian History & \numOne{229} & 4 & \numOne{14.048034934497817} & \numOne{2.8013100436681224} & $2,287$ ($10,620$) \\
        PzHistory & 183 & 3 & \numOne{38.885245901639344} & \numOne{2.4353369763205825} & $1,261$ ($7,518$) \\
        \hline
        Overall & $2,633$ & \numOne{3.93} & \numOne{15.666160849772382} & \numOne{2.8868835054531417} & $13,329$ ($56,104$)\\
        \hline
        \hline
        \multicolumn{6}{c}{RACE Train - Mid and High School} \\
        \hline
        RACE-M & $25,421$ & \numOne{4} & \numOne{9.0} & \numOne{3.9} &  $32,811$ \\
        RACE-H & $62,445$ & \numOne{4} & \numOne{10.4} & \numOne{5.8} & $125,120$ \\
        \hline
        Overall & $87,866$ & \numOne{4} & \numOne{10.0} & \numOne{5.3} & $136,629$ \\
        \bottomrule
    \end{tabular}
    \caption{Statistics about our Bulgarian dataset compared to the RACE dataset.}
    \label{tab:data_stats}
    \vspace{-0.2in}
\end{table*}{}

While there exist many datasets for reading comprehension, most of them are in English, and there are a very limited number in other languages \citep{Penas_overviewof,Penas_overviewof2014}. Hereby, we collect our own dataset for Bulgarian, resulting in 2,633 multiple-choice questions, without contexts, from different subjects: biology~(16.6\%), philosophy~(23.93\%), geography~(23.24\%), and history~(36.23\%). 
Table~\ref{tab:sample_data} shows an example question with candidate answers chosen to represent best each category. We use green to mark the correct answer, and bold for the question category. For convenience all the examples are translated to English.

\begin{table}[htbp]
            \small{
                (\textbf{Biology}) 
                The thick coat of mammals in winter is an example of: \\
                A. physiological adaptation \\
                B. behavioral adaptation \\
                C. genetic adaptation \\
                D. \textcolor{lincolngreen}{morphological adaptation}
                \medskip
                
                (\textbf{Philosophy}) 
                According to relativism in ethics: \\
                A. there is only one moral law that is valid for all \\
                B. \textcolor{lincolngreen}{there is no absolute good and evil} \\
                C. people are evil by nature  \\
                D. there is only good, and the evil is seeming
                \medskip
                
                (\textbf{Geography}) 
                Which of the assertions about the economic specialization of the Southwest region is true? \\
                A. The ratio between industrial and agricultural production is 15:75 \\
                B. \textcolor{lincolngreen}{Lakes of glacial origin in Rila and Pirin are a resource for the development of tourism} \\
                C. Agricultural specialization is related to the cultivation of grain and ethereal-oil crops \\
                D. The rail transport is of major importance for intra-regional connections 
                \medskip
                
                (\textbf{History}) 
                Point out the concept that is missed in the text of the Turnovo Constitution: ,,Art. 54 All born in Bulgaria, also those born elsewhere by parents Bulgarian \underline{\hspace{1cm}}, count as \underline{\hspace{1cm}}  of the Bulgarian Principality. Art. 78 Initial teaching is free and obligatory for all \underline{\hspace{1cm}} of the Bulgarian Principality.''\\
                A. residents \\
                B. \textcolor{lincolngreen}{citizents} \\
                C. electors \\
                D. voters 
                \medskip
                
                (\textbf{History Quiz}) 
                Sofroniy Vrachanski started a family that plays a big role in the history of the Bulgarian National Revival. What is its name? \\
                A. Georgievi \\
                B. Tapchileshtovi \\
                C. \textcolor{lincolngreen}{Bogoridi} \\
                D. Palauzovi
            }%

    \normalsize
    \caption{Example questions, one per subject, from our Bulgarian dataset. The correct answer is marked in green.}
    \label{tab:sample_data}
\end{table}{}

Table~\ref{tab:data_stats} shows the distribution of questions per subject category, the length (in words) for both the questions and the options (candidate answers), and the vocabulary richness, measured in terms of unique words. The first part of the table presents statistics about our dataset, while the second part is a comparison to RACE~\citep{lai-etal-2017-race}. 

We divided the Bulgarian questions into two groups based on the question's source. The first group (\textit{12th Grade Matriculation Exam}) was collected from twelfth grade matriculation exams created by the Ministry of Education of Bulgaria in the period 2008--2019. Each exam contains thirty multiple-choice questions with four possible answers per question. The second set of questions (\textit{Online History Quizzes}) are history-related and are collected from online quizzes. While they are not created by educators, the questions are still challenging and well formulated. Furthermore, we manually filtered out questions with non-textual content (i.e.,~pictures, paintings, drawings, etc.), ordering questions (i.e.,~order the historical events), and questions involving calculations (i.e.,~how much $X$ we need to add to $Y$ to arrive at $Z$).

Table~\ref{tab:data_stats} shows that history questions in general contain more words (\numOne{14.0}--\numOne{38.9} on average), compared to other subjects (\numOne{8.9}--\numOne{12.8} on average). A tangible difference in length compared to other subjects is seen for \textit{12th grade History} and \textit{PzHistory}, due to the large number of quotes, and document pieces contained in questions from these two groups. Also, the average question length is \numOne{15.666160849772382}, which is longer compared to the RACE dataset with \numOne{10.0}. On the other hand, the option lengths per subject category in our dataset follow a narrower distribution. They fall in the interval between \numOne{2.5} and \numOne{2.9} words on average, expect for \textit{12th grade History}, with \numOne{3.6} words. Here, we note a significant difference compared to the option lengths in RACE, which tend to be \numOne{2.4} words longer on average -- \numOne{5.3} for RACE vs. \numOne{2.9} for ours.

\noindent Finally, we examine the vocabulary richness of the two datasets. The total number of unique words is shown in the last column of Table~\ref{tab:data_stats} (Vocab Size). For our dataset, there are two numbers per row: the first one shows statistics based on the question--answer pairs only, while the second one, enclosed in parentheses, measures the vocabulary size including the extracted passages by the Context Retriever. The latter number is a magnitude estimate rather then a concrete number, since its upper limit is the number of words in Wikipedia, and it can vary for different retrieval strategies. 

\section{Experiments and Evaluation}
\label{sec:experiments}

\subsection{BERT Fine-Tuning}
We divide the fine-tuning into two groups of models (\textit{i})~Multilingual BERT, and (\textit{ii})~Slavic BERT. Table~\ref{tab:race_results} below presents the results in the multiple-choice comprehension task on the dev dataset from RACE~\cite{lai-etal-2017-race}.

\begin{table}[h]
    \centering
    \begin{tabular}{cccc}
        \toprule
         \bf \#Epoch & \bf RACE-M & \bf RACE-H & \bf Overall  \\
         \midrule
         BERT 1 & \numTwo{64.20612813370473} & \numTwo{53.65923384791309} & \numTwo{56.72882042967167} \\
         BERT 2 & \numTwo{68.80222841225627} & \numTwo{57.57575757575758}  & \numTwo{60.84312930685043}  \\
         BERT 3 & \numTwo{69.15041782729805} & \numTwo{58.4333905088622}  & \numTwo{61.55249290636401}  \\
         \hline
         Slavic 2 & \numTwo{53.55153203342619} & \numTwo{44.48256146369354} & \numTwo{47.12201053911634} \\
         Slavic 3 & \numTwo{57.38161559888579} & \numTwo{46.88393367638651} & \numTwo{49.93919740575598} \\
         \bottomrule
    \end{tabular}
    \caption{Accuracy measured on the dev RACE dataset after each training epoch.}
    \label{tab:race_results}
    \vspace{-0.2in}
\end{table}{}

\label{sec:bertfinetune}

\paragraph{Multilingual BERT} 
As our initial model, we use BERT\textsubscript{base}, Multilingual Cased
which is pre-trained on 104 languages, and has 12-layers, 768-hidden units per layer, 12-heads, and a total of 110M parameters. We further fine-tune the model on RACE~\cite{lai-etal-2017-race} for 3 epochs saving a checkpoint after each epoch. We use a batch size of 8, a max sequence size of 320, and a learning rate of 1e-5.

\paragraph{Slavic BERT} 
The Slavic model\footnote{\url{http://github.com/deepmipt/Slavic-BERT-NER}} was built using transfer learning from the Multilingual BERT model to four Slavic languages: Bulgarian, Czech, Polish, and Russian. In particular, the Multilingual BERT model was fine-tuned on a stratified dataset of Russian news and Wikipedia articles for the other languages. We use this pre-trained Slavic BERT model, and we apply the same learning procedure as for \textit{Multilingual BERT}.

\subsection{Wikipedia Retrieval and Indexing}

Here, we discuss the retrieval setup (see Section~\ref{sec:ctxretriever} for more details). We use the Bulgarian dump of Wikipedia from 2019-04-20, with a total of 251,507 articles. We index each article title and body in plain text, which we call a \emph{passage}. We further apply additional processing for each field:

\begin{itemize}[noitemsep,topsep=2pt,parsep=2pt,partopsep=2pt]
    \item \textit{ngram}: word-based 1--3 grams;
    \item \textit{bg}: lowercased, stop-words removed (from Lucene), and stemmed~\citep{savoy2007searching};
    \item \textit{none}: bag-of-words index.
\end{itemize}{}

We ended up using a subset of four fields from all the possible analyzer-field combinations, namely \textit{title.bg, passage, passage.bg,} and \textit{passage.ngram}. We applied Bulgarian analysis on the \emph{title} field only as it tends to be short and descriptive, and thus very sensitive to noise from stop-words, which is in contrast to questions that are formed mostly of stop-words, e.g., \emph{what}, \emph{where}, \emph{when}, \emph{how}.

For indexing the Wikipedia articles, we adopt two strategies: sliding window and paragraph. In the window-based strategy, we define two types of splits: small, containing 80-100 words, and large, of around 300 words. In order to obtain indexing chunks, we define a window of size $K$, and a stride equal to one forth of $K$. Hence, each $\frac{K}{4}$ characters, which is the size of the stride, are contained into four different documents. The paragraph-based strategy divides the article by splitting it using one or more successive newline characters ({[\textbackslash{n}]+}) as a delimiter. We avoid indexing entire documents due to their extensive length, which can be far beyond the maximum length that BERT can take as an input, i.e., 320 word pieces (see Section~\ref{sec:bertfinetune} for the more details). Note that extra steps are needed in order to extract a proper passage from the text. Moreover, the amount of facts in the Wikipedia articles that are unrelated to our questions give rise to false positives since the question is short and term-unspecific.

Finally, we use a list of top-$N$ hits for each candidate answer. Thus, we have to execute an additional query for each question + option combination, which may result in duplicated passages, thus introducing an implicit bias towards the candidates they support. In order to mitigate this effect, during the answer selection phase (see Section~\ref{sec:selstrategies}), we remove all duplicate entries, keeping a single instance.

\subsection{Experimental Results}
\label{sec:expres}

Here, we discuss the accuracy of each model on the original English MRC task, followed by experiments in zero-shot transfer to Bulgarian.

\paragraph{English Pre-training for MCRC.}
Table~\ref{tab:race_results} presents the change in accuracy on the original English comprehension task, depending on the number of training epochs. In the table, ``BERT'' refers to the Multilingual BERT model, while ``Slavic'' stands for BERT with Slavic pre-training. We further fine-tune the models on the RACE dataset. Next, we report their performance in terms of accuracy, following the notation from \citep{lai-etal-2017-race}. Note that the questions in RACE-H are more complex than those in RACE-M. The latter has more word matching questions and fewer reasoning questions. The final column in the table, \textit{Overall}, shows the accuracy calculated over all questions in the RACE testset. We train both setups for three epochs and we report their performance after each epoch. We can see a positive correlation between the number of epochs and the model's accuracy. We further see that the Slavic BERT performs far worse on both RACE-M and RACE-H, which suggests that the change of weights of the model towards Slavic languages has led to catastrophic forgetting of the learned English syntax and semantics. Thus, it should be expected that the adaptation to Slavic languages would yield decrease in performance for English. What matters though is whether this helps when testing on Bulgarian, which we explore next.

\begin{table}[t]
    \centering
    \begin{tabular}{lr}
        \toprule
         \bf Setting & \bf Accuracy \\
         \midrule
         Random & 24.89 \\
         \hline
         Train for 3 epochs & -- \\
         + window \& title.bg \& pass.ngram & \numTwo{29.62} \\
         + passage.bg \& passage & \numTwo{39.35} \\
         -- title.bg & \numTwo{39.69} \\
         + passage.bg\textasciicircum 2 & \numTwo{40.26} \\
         + title.bg\textasciicircum 2  & \numTwo{40.30} \\
         \hline
         + bigger window & \numTwo{36.54} \\
         + paragraph split & \numTwo{42.23} \\
         \hline
         + Slavic pre-training & \numTwo{33.270034181541966} \\
         \hline
         Train for 1 epoch best & \numTwo{40.25826053930877} \\
         Train for 2 epochs best & \numTwo{41.89137865552602} \\
         \bottomrule
    \end{tabular}
    \caption{Accuracy on the Bulgarian testset: ablation study when sequentially adding/removing different model components.}
    \label{tab:model_res_short}
    \vspace{-0.1in}
\end{table}{}

\paragraph{Zero-Shot Transfer.}
Here, we assess the performance of our model when applied to Bulgarian multiple-choice reading comprehension. Table~\ref{tab:model_res_short} presents an ablation study for various components. Each line denotes the type of the model, and the addition (+) or the removal \mbox{(--)} of a characteristic from the setup in the previous line. The first line shows the performance of a baseline model that chooses an option uniformly at random from the list of candidate answers for the target question. The following rows show the results for experiments conducted with a model trained for three epochs on RACE~\citep{lai-etal-2017-race}. 

Our basic model uses the following setup: Wikipedia pages indexed using a small sliding window (400 characters, and stride of 100 characters), and context retrieval over two fields: Bulgarian analyzed title (\textit{text.bg}), and word $n$-grams over the passage (\textit{passage.ngram}). This setup yields 29.62\% accuracy, and it improves over the random baseline by 4.73\% absolute. We can think of it as a non-random baseline for further experiments. 
Next, we add two more fields to the IR query: passage represented as a bag of words (named \textit{passage}), and Bulgarian analyzed (\textit{passage.bg}), which improves the accuracy by additional 10\%, arriving at 39.35\%. The following experiment shows that removing the \textit{title.bg} field does not change the overall accuracy, which makes it an insignificant field for searching. Further, we add double weight on \textit{passage.bg}, (shown as \textasciicircum{2}), which yields 1\% absolute improvement.

From the experiments described above, we found the best combination of query fields to be \textit{title.bulgarian\textasciicircum{2}, passage.ngram, passage, passage.bulgarian\textasciicircum{2}}, where the \textit{title} has a minor contribution, and can be sacrificed for ease of computations and storage. Fixing the best query fields, allowed us to evaluate other indexing strategies, i.e.,~bigger window (size 1,600, stride 400) with accuracy 36.54\%, and paragraph splitting, with which we achieved our highest accuracy of 42.23\%. This is an improvement of almost 2.0\% absolute over the small sliding window, and 5.7\% over the large one.

Next, we examined the impact of the Slavic BERT. Surprisingly, it yielded 9\% absolute drop in accuracy compared to the multi-lingual BERT. This suggests that the latter already has enough knowledge about Bulgarian, and thus it does not need further adaptation to Slavic languages.

\begin{figure}[tbh]
    \centering
    \includegraphics[width=\columnwidth]{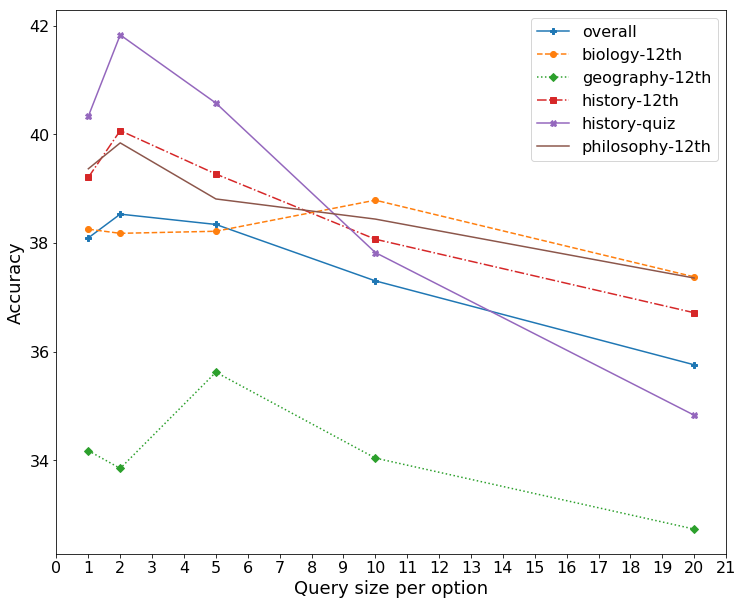}
    \caption{Accuracy per question category based on the number of query results per answer option.}
    \label{fig:acc_per_cat}
\end{figure}{}

Next, we study the impact of the number of fine-tuning epochs on the model's performance. We observe an increase in accuracy as the number of epochs grows, which is in line with previously reported results for English tasks. While this correlation is not as strong as for the original RACE task (see Table~\ref{tab:race_results} for comparison), we still observe 1.6\% and 0.34\% absolute increase in accuracy for epochs 2 and 3, respectively, compared to epoch~1. Note that we do not go beyond three epochs, as previous work has suggested that 2-3 fine-tuning epochs are enough \cite{devlin2018bert}, and after that, there is a risk of catastrophic forgetting of what was learned at pre-training time (note that we have already seen such forgetting with the Slavic BERT above).

We further study the impact of the size  of the results list returned by the retriever on the accuracy for the different categories. Figure~\ref{fig:acc_per_cat} shows the average accuracy for a given query size $S_q$ over all performed experiments, where $S_q~\in~{\{1, 2, 5, 10, 20\}}$. 
We can see in Figure~\ref{fig:acc_per_cat} that longer query result lists (i.e.,~containing more than 10 results) per answer option worsen the accuracy for all categories, except for \emph{biology}, where we see a small peak at length 10, while still the best overall results for this category is achieved for a result list of length 5. A single well-formed maximum at length 2 is visible for \emph{history} and \emph{philosophy}. With these two categories being the biggest ones, the cap at the same number of queries for the overall accuracy is not a surprise. 
The per-category results for the experiments are discussed in more detail in Appendix~\ref{sec:results_per_cat}. 

\noindent We can see that the highest accuracy is observed for \emph{history}, particularly for online quizzes, which are not designed by educators and are more of a word-matching nature rather then a reasoning one (see Table~\ref{tab:sample_data}). 
Finally, \emph{geography} appears to be the hardest category with only 38.73\% accuracy: 3.5\% absolute difference compared to the second-worst category. The performance for this subject is also affected differently by changes in query result length: the peak is at lengths 5 and 10, while there is a drop for length 2. A further study of the model's behavior can be found in Appendix~\ref{sec:contextretbert}.
\section{Conclusion and Future Work}
\label{sec:conclustions}

We studied the task of multiple-choice reading comprehension for low-resource languages, using a newly collected Bulgarian corpus with 2,633 questions from matriculation exams for twelfth grade in history and biology, and online exams in history without explanatory contexts. In particular, we designed an end-to-end approach, on top of a multilingual BERT model \citep{devlin2018bert}, which we fine-tuned on large-scale English reading comprehension corpora, and open-domain commonsense knowledge sources (Wikipedia). Our main experiments evaluated the model when applied to Bulgarian in a zero-shot fashion. The experimental results found additional pre-training on the English RACE corpus to be very helpful, while pre-training on Slavic languages to be harmful, possibly due to catastrophic forgetting. Paragraph splitting, $n$-grams, stop-word removal, and stemming further helped the context retriever to find better evidence passages, and the overall model to achieve accuracy of up to \numTwo{42.23319407519939}\%, which is well above the baselines of 24.89\% and 29.62\%.

In future work, we plan to make use of reading strategies~\citep{sun-etal-2019-improving}, linked entities~\citep{DBLP:journals/corr/abs-1902-00993}, concatenation and reformulation of passages and questions~\citep{simov2012bulgarian,clark2016combining,ni-etal-2019-learning}, as well as re-ranking of documents~\citep{nogueira2019passage}.

\section*{Acknowledgments}

We want to thank Desislava Tsvetkova and Anton Petkov for the useful discussions and for their help with some of the experiments.

This research is partially supported by Project UNITe BG05M2OP001-1.001-0004 funded by the OP ``Science and Education for Smart Growth'' and co-funded by the EU through the ESI Funds.

\bibliographystyle{acl_natbib}
\bibliography{acl2017}

\appendix

\section*{Appendix}
\section{Per-Category Results}
\label{sec:results_per_cat}

Table~\ref{tab:model_results} gives an overview, including per-category breakdown, of our parameter tuning experiments. We present the results for some interesting experiments rather then for a full grid search. The first row shows a random baseline for each category. In the following rows, we compare different types of indexing: first, we show the results for a small sliding window (400-character window, and 100-character stride), followed by a big window (1,600-character window, and 400-character stride), and finally for paragraph indexing. We use the same notation as in Section~\ref{sec:experiments}. The last group in the table (\textit{Paragraph}) shows the best-performing model, where we mark in bold the highest accuracy for each category. For completeness, we also show the accuracy when using the \textit{Slavic BERT} model for prediction, which yields a 10\% drop on average compared to using the \textit{Multilingual BERT}, for each of the categories. 

\begin{table*}[t]
    \centering
    \small
    \begin{tabular}{cccccccc}
    \toprule
    
    \bf \#docs & \bf Overall & \bf biology-12th & \bf philosophy-12th & \bf geography-12th & \bf history-12th & \bf history-quiz \\

\midrule
\multicolumn{7}{c}{Random} \\
\hline
0 & \numTwo{24.88619119878604} & \numTwo{26.08695652173913} & \numTwo{24.444444444444443} & \numTwo{24.18300653594771} & \numTwo{25.871559633027523} & \numTwo{24.02912621359223} \\
\hline
\hline
\multicolumn{7}{c}{Window Small} \\
\hline
\multicolumn{7}{c}{title.bulgarian, passage.bulgarian} \\
\hline
1 & \numTwo{39.95442461071022} & \numTwo{40.274599542334094} & \numTwo{40.63492063492063} & \numTwo{34.967320261437905} & \numTwo{42.988929889298895} & \numTwo{41.99029126213592} \\
2 & \numTwo{40.22028104823395} & \numTwo{40.274599542334094} & \numTwo{40.63492063492063} & \numTwo{35.947712418300654} & \numTwo{42.61992619926199} & \numTwo{42.71844660194175} \\
5 & \numTwo{40.22028104823395} & \numTwo{38.901601830663616} & \numTwo{40.63492063492063} & \numTwo{38.071895424836605} & \numTwo{41.51291512915129} & \numTwo{42.47572815533981} \\
10 & \numTwo{38.663121914166354} & \numTwo{40.503432494279174} & \numTwo{39.84126984126984} & \numTwo{35.45751633986928} & \numTwo{39.298892988929886} & \numTwo{38.83495145631068} \\
20 & \numTwo{36.84010634257501} & \numTwo{37.52860411899314} & \numTwo{39.04761904761905} & \numTwo{33.8235294117647} & \numTwo{38.745387453874535} & \numTwo{34.70873786407767} \\
\hline
\multicolumn{7}{c}{title.bulgarian, passage.ngram} \\
\hline
1 & \numTwo{28.940372199012533} & \numTwo{29.06178489702517} & \numTwo{32.06349206349206} & \numTwo{27.287581699346404} & \numTwo{27.490774907749078} & \numTwo{28.398058252427184} \\
2 & \numTwo{29.092290163311812} & \numTwo{29.06178489702517} & \numTwo{33.333333333333336} & \numTwo{25.0} & \numTwo{28.782287822878228} & \numTwo{29.12621359223301} \\
5 & \numTwo{29.05431067223699} & \numTwo{27.45995423340961} & \numTwo{32.06349206349206} & \numTwo{26.633986928104576} & \numTwo{30.627306273062732} & \numTwo{27.66990291262136} \\
10 & \numTwo{29.624003038359287} & \numTwo{29.06178489702517} & \numTwo{32.53968253968254} & \numTwo{26.96078431372549} & \numTwo{30.07380073800738} & \numTwo{29.12621359223301} \\
20 & \numTwo{29.43410558298519} & \numTwo{31.80778032036613} & \numTwo{32.698412698412696} & \numTwo{26.633986928104576} & \numTwo{28.59778597785978} & \numTwo{27.184466019417474} \\
\hline
\multicolumn{7}{c}{title.bulgarian, passage.ngram, passage, passage.bulgarian} \\
\hline
1 & \numTwo{38.32130649449297} & \numTwo{38.215102974828376} & \numTwo{40.0} & \numTwo{34.47712418300654} & \numTwo{39.48339483394834} & \numTwo{40.04854368932039} \\
2 & \numTwo{39.080896315989364} & \numTwo{37.07093821510298} & \numTwo{40.317460317460316} & \numTwo{34.47712418300654} & \numTwo{40.59040590405904} & \numTwo{44.1747572815534} \\
5 & \numTwo{39.3467527535131} & \numTwo{40.961098398169334} & \numTwo{39.84126984126984} & \numTwo{34.64052287581699} & \numTwo{41.32841328413284} & \numTwo{41.262135922330096} \\
10 & \numTwo{38.62514242309153} & \numTwo{40.503432494279174} & \numTwo{40.63492063492063} & \numTwo{33.49673202614379} & \numTwo{40.40590405904059} & \numTwo{38.83495145631068} \\
20 & \numTwo{36.53627041397645} & \numTwo{38.672768878718536} & \numTwo{37.93650793650794} & \numTwo{31.372549019607842} & \numTwo{37.45387453874539} & \numTwo{38.592233009708735} \\
\hline
\multicolumn{7}{c}{passage.ngram, passage, passage.bulgarian\textasciicircum{2}} \\
\hline
1 & \numTwo{39.68856817318648} & \numTwo{40.274599542334094} & \numTwo{40.63492063492063} & \numTwo{35.130718954248366} & \numTwo{42.06642066420664} & \numTwo{41.262135922330096} \\
2 & \numTwo{40.25826053930877} & \numTwo{39.816933638443935} & \numTwo{40.95238095238095} & \numTwo{35.947712418300654} & \numTwo{42.61992619926199} & \numTwo{42.96116504854369} \\
5 & \numTwo{39.57462969996202} & \numTwo{39.588100686498855} & \numTwo{39.36507936507937} & \numTwo{37.254901960784316} & \numTwo{40.95940959409594} & \numTwo{41.50485436893204} \\
10 & \numTwo{38.70110140524117} & \numTwo{41.18993135011441} & \numTwo{39.523809523809526} & \numTwo{35.78431372549019} & \numTwo{39.298892988929886} & \numTwo{38.349514563106794} \\
20 & \numTwo{37.143942271173565} & \numTwo{39.359267734553775} & \numTwo{37.77777777777778} & \numTwo{35.294117647058826} & \numTwo{38.37638376383764} & \numTwo{34.95145631067961} \\
\hline
\multicolumn{7}{c}{title.bulgarian\textasciicircum{2}, passage.ngram, passage, passage.bulgarian\textasciicircum{2}} \\
\hline
1 & \numTwo{39.840486137485755} & \numTwo{40.274599542334094} & \numTwo{40.79365079365079} & \numTwo{35.130718954248366} & \numTwo{42.25092250922509} & \numTwo{41.74757281553398} \\
2 & \numTwo{40.29624003038359} & \numTwo{40.274599542334094} & \numTwo{40.63492063492063} & \numTwo{36.111111111111114} & \numTwo{42.80442804428044} & \numTwo{42.71844660194175} \\
5 & \numTwo{40.25826053930877} & \numTwo{39.130434782608695} & \numTwo{40.63492063492063} & \numTwo{38.39869281045752} & \numTwo{41.14391143911439} & \numTwo{42.47572815533981} \\
10 & \numTwo{38.73908089631599} & \numTwo{40.503432494279174} & \numTwo{39.682539682539684} & \numTwo{35.62091503267974} & \numTwo{39.48339483394834} & \numTwo{39.077669902912625} \\
20 & \numTwo{37.06798328902393} & \numTwo{37.75743707093822} & \numTwo{39.04761904761905} & \numTwo{34.64052287581699} & \numTwo{38.56088560885609} & \numTwo{34.95145631067961} \\
\hline
\hline
\multicolumn{7}{c}{Window Big} \\
\hline
\multicolumn{7}{c}{title.bulgarian\textasciicircum{2}, passage.ngram, passage, passage.bulgarian\textasciicircum{2}} \\
\hline
1 & \numTwo{31.21914166350171} & \numTwo{28.37528604118993} & \numTwo{33.96825396825397} & \numTwo{29.41176470588235} & \numTwo{30.81180811808118} & \numTwo{33.25242718446602} \\
2 & \numTwo{33.11811621724269} & \numTwo{31.57894736842105} & \numTwo{37.46031746031746} & \numTwo{31.209150326797385} & \numTwo{33.94833948339483} & \numTwo{29.854368932038835} \\
5 & \numTwo{36.0425370300038} & \numTwo{35.69794050343249} & \numTwo{38.095238095238095} & \numTwo{33.8235294117647} & \numTwo{37.82287822878229} & \numTwo{34.22330097087379} \\
10 & \numTwo{36.53627041397645} & \numTwo{37.29977116704806} & \numTwo{36.03174603174603} & \numTwo{33.98692810457516} & \numTwo{39.298892988929886} & \numTwo{36.650485436893206} \\
20 & \numTwo{35.62476262818078} & \numTwo{34.55377574370709} & \numTwo{39.682539682539684} & \numTwo{31.045751633986928} & \numTwo{38.37638376383764} & \numTwo{33.737864077669904} \\
\hline
\hline
\multicolumn{7}{c}{Paragraph} \\
\hline
\multicolumn{7}{c}{title.bulgarian\textasciicircum{2}, passage.ngram, passage, passage.bulgarian\textasciicircum{2}} \\
\hline
1 & \numTwo{41.815419673376375} & \numTwo{41.41876430205949} & \numTwo{42.06349206349206} & \numTwo{38.071895424836605} & \numTwo{40.95940959409594} & \numTwo{48.54368932038835} \\
2 & \textbf{\numTwo{42.23319407519939}} & \numTwo{42.5629290617849} & \textbf{\numTwo{43.17460317460318}} & \numTwo{35.62091503267974} & \textbf{\numTwo{42.988929889298895}} & \textbf{\numTwo{49.271844660194176}} \\
5 & \numTwo{41.58754272692746} & \textbf{\numTwo{43.24942791762014}} & \numTwo{40.317460317460316} & \textbf{\numTwo{38.72549019607843}} & \numTwo{40.03690036900369} & \numTwo{48.05825242718446} \\
10 & \numTwo{39.46069122673756} & \numTwo{40.961098398169334} & \numTwo{38.41269841269841} & \numTwo{36.928104575163395} & \numTwo{39.85239852398524} & \numTwo{42.71844660194175} \\
20 & \numTwo{37.52373718192176} & \numTwo{39.130434782608695} & \numTwo{37.61904761904762} & \numTwo{34.64052287581699} & \numTwo{38.56088560885609} & \numTwo{38.592233009708735} \\
\hline
\multicolumn{7}{c}{Slavic BERT} \\
\hline
1 & \numTwo{33.19407519939233} & \numTwo{30.892448512585812} & \numTwo{33.17460317460318} & \numTwo{28.758169934640524} & \numTwo{32.28782287822878} & \numTwo{43.44660194174757} \\
2 & \numTwo{33.270034181541966} & \numTwo{31.57894736842105} & \numTwo{31.904761904761905} & \numTwo{31.209150326797385} & \numTwo{35.239852398523986} & \numTwo{37.62135922330097} \\
5 & \numTwo{31.14318268135207} & \numTwo{30.205949656750573} & \numTwo{30.158730158730158} & \numTwo{29.248366013071895} & \numTwo{30.99630996309963} & \numTwo{36.650485436893206} \\
10 & \numTwo{30.421572350930496} & \numTwo{29.290617848970253} & \numTwo{29.682539682539684} & \numTwo{29.73856209150327} & \numTwo{31.918819188191883} & \numTwo{31.796116504854368} \\
20 & \numTwo{29.661982529434106} & \numTwo{28.60411899313501} & \numTwo{29.365079365079364} & \numTwo{28.431372549019606} & \numTwo{32.103321033210335} & \numTwo{29.854368932038835} \\
\bottomrule
    \end{tabular}
    \caption{Evaluation results for the Bulgarian multiple-choice reading comprehension task: comparison of various indexing and query strategies.}
    \label{tab:model_results}
\end{table*}{}

\section{Case Study}
\label{sec:contextretbert}

In Table~\ref{tab:context_pred}, we present the retrieved evidence passages for the example questions in Table~\ref{tab:sample_data}: we omit the answers, and we only show the questions and the contexts. Each example is separated by a double horizontal line, where the first row is the question starting with ``\textit{Q:}'', and the following rows contain passages returned by the retriever. For each context, we normalize the raw scores from the comprehension model using  Eq.~\ref{eq:bertsoftmax} to obtain a probability distribution. 
We then select an answer using $\argmax$, according to Eq.~\ref{eq:anssel}. In the table, we indicate the correctness of each predicted answer using one of the following symbols before the question:
\begin{itemize}
    \item[\correctmark] The question is answered correctly.
    \item[\incorrectmark] An incorrect answer has the highest score.
    \item[\questionmark] Two or more answers have the highest score.
\end{itemize}{}

We show the top retrieved result in order to illustrate the model scores over different evidence passages and the quality of the articles. The queries are formed by concatenating the question with an answer option, even though this can lead to duplicate results since some answers can be quite similar or the question's terms could dominate the similarity score.

\noindent The questions in Table~\ref{tab:context_pred} are from five different categories: biology, philosophy, geography, history, and online quizzes. Each of them has its own specifics and gives us an opportunity to illustrate a different model behavior.

The first question is from the biology domain, and we can see that the text is very general, and so is the retrieved context. The latter talks about \emph{hair} rather than \emph{coat}, and the correct answer (D) \emph{morphological adaptation} is not present in the retrieved text. On the other hand, all the terms are only connected to it, and hence the model assigns high probability to this answer option.

For the second question, from the philosophy domain, there are two related contexts found. The first one is quite short, noisy, and it does not give much information in general. The second paragraph manages to extract the definition of \emph{relativism} and to give good supporting evidence for the correct answer, namely that \emph{there is no absolute good and evil} (B). As a result, this option is assigned high probability. Nevertheless, the incorrect answer \emph{here is only one moral law that is valid for all} (A) is assigned an even higher probability and it wins the voting.

In the third example, from the domain of geography, we see a large number of possible contexts, due to the long and descriptive answers. We can make two key observations: (\textit{i})~the query is drawn in very different directions by the answers, and (\textit{ii})~there is no context for \emph{Southwestern region}, and thus,
in the second option, the result is for Russia, not for Bulgaria. The latter passage pushes the probability mass to an option that talks about transportation (D), which is incorrect. Fortunately, the forth context has an almost full term overlap with the correct answer (B), and thus gets very high probability assigned to it:  72\%.

The fourth question, from the history domain, asks to point out a missing concept, but the query is dominated by the question, and especially by underscores, leading to a single hit, counting only symbols, without any words. As expected, the model assigned uniform probability to all classes.

The last question, a history quiz, is a factoid one, and it lacks a reasoning component, unlike the previous examples. The query returned a single direct match. The retrieved passage contains the correct answer exactly: option \emph{Bogoridi} (C). Thereby, the comprehension model assigns to it a very high probability of 68\%.

\begin{table*}[t]
    \begin{tabularx}{\textwidth}{Xllll}
        \toprule
         \bf Context & $Pr_A$ & $Pr_B$  & $Pr_C$  & $Pr_D$  \\
         \midrule
         {\correctmark} Q: The thick coat of mammals in winter is an example of:  & & & &  \\
         1) The hair cover is a rare and rough bristle. In winter, soft and dense hair develops between them. Color ranges from dark brown to gray, individually and geographically diverse & \numTwo{0.19} & \numTwo{0.19} & \numTwo{0.15} & \numTwo{0.47} \\
         \hline
         \hline
         {\incorrectmark} Q: According to relativism in ethics: & & & & \\
         1) Moral relativism & \numTwo{0.45} & \numTwo{0.24} & \numTwo{0.10} & \numTwo{0.21} \\
         2) In ethics, relativism is opposed to absolutism. Whilst absolutism asserts the belief that there are universal ethical standards that are inflexible and absolute, relativism claims that ethical norms vary and differ from age to age and in different cultures and situations. It can also be called epistemological relativism - a denial of absolute standards of truth evaluation. & \numTwo{0.28} & \numTwo{0.41} & \numTwo{0.09} & \numTwo{0.22} \\
         \hline
         \hline
         {\correctmark} Q: Which of the assertions about the economic specialization of the Southwest region is true? & & & & \\
         1) Geographic and soil-climatic conditions are blessed for the development and cultivation of oil-bearing rose and other essential oil crops. & \numTwo{0.12} & \numTwo{0.52} & \numTwo{0.28} & \numTwo{0.08} \\
         2) Kirov has an airport of regional importance. Kirov is connected with rail transport with the cities of the Transsiberian highway (Moscow and Vladivostok). & \numTwo{0.14} & \numTwo{0.27} & \numTwo{0.06} & \numTwo{0.53} \\ 
         3) Dulovo has always been and remains the center of an agricultural area, famous for its grain production. The industrial sectors that still find their way into the city's economy are primarily related to the primary processing of agricultural produce. There is also the seamless production that evolved into small businesses with relatively limited economic significance. & \numTwo{0.25} & \numTwo{0.05} & \numTwo{0.67} & \numTwo{0.03} \\
         4) In the glacial valleys and cirques and around the lakes in the highlands of Rila and Pirin, there are marshes and narrow-range glaciers (overlaps). & \numTwo{0.10} & \numTwo{0.72} & \numTwo{0.08} & \numTwo{0.10} \\
         \hline
         \hline
         {\questionmark} Q: Point out the concept that is missed in the text of the Turnovo Constitution: \dots & & & &\\
         1) \underline{\hspace{5cm}} & \numTwo{0.26} & \numTwo{0.26} & \numTwo{0.26} & \numTwo{0.22} \\
         \hline
         \hline
         {\correctmark} Q: Sofroniy Vrachanski sets up a genre that plays a big role in the history of the Bulgarian Revival. What is his name? & & & & \\
         1) Bogoridi is a Bulgarian Chorbadji genus from Kotel. Its founder is Bishop Sofronius Vrachanski (1739-1813). His descendants are: & 
         \numTwo{0.06} & \numTwo{0.16} & \numTwo{0.68} & \numTwo{0.10} \\
         \bottomrule
         
    \end{tabularx}{}
    \caption{Retrieved unique top-1 contexts for the example questions in Table~\ref{tab:sample_data}. The passages are retrieved using queries formed by concatenating a question with an answer option.}
    \label{tab:context_pred}
\end{table*}{}

\end{document}